\documentclass[conference]{IEEEtran}
\IEEEoverridecommandlockouts
\usepackage{cite}
\usepackage{amsmath,amssymb,amsfonts}
\usepackage{algorithmic}
\usepackage{graphicx}
\usepackage{textcomp}
\usepackage{multirow}
\newcommand{\RNum}[1]{\uppercase\expandafter{\romannumeral #1\relax}}
\usepackage{xcolor}
\DeclareRobustCommand*{\IEEEauthorrefmark}[1]{%
    \raisebox{0pt}[0pt][0pt]{\textsuperscript{\footnotesize\ensuremath{#1}}}}
\def\BibTeX{{\rm B\kern-.05em{\sc i\kern-.025em b}\kern-.08em
    T\kern-.1667em\lower.7ex\hbox{E}\kern-.125emX}}

\title{Distribution Context Aware Loss for Person Re-identification}
\author{
    \IEEEauthorblockN{Zhigang Chang\IEEEauthorrefmark{1}, 
    Qin Zhou\IEEEauthorrefmark{2}, 
    Mingyang Yu\IEEEauthorrefmark{1}, 
    Shibao Zheng\IEEEauthorrefmark{1},
    Hua Yang\IEEEauthorrefmark{1}
    Tai Pang Wu\IEEEauthorrefmark{3}}
    \IEEEauthorblockA{\IEEEauthorrefmark{1}Institute of Image Processing and Network Engineering, Shanghai Jiao Tong University,  Shanghai 200240, China}
    \IEEEauthorblockA{\IEEEauthorrefmark{2}Artificial Intelligence Center-City Brain, Alibaba Cloud, Hangzhou 311100, China}
    \IEEEauthorblockA{\IEEEauthorrefmark{3}1000 Video Technology Co. Limited, Suzhou, China
    \\\{changzig,oooofish, sbzh, hyang\}@sjtu.edu.cn,
        xining.zq@alibaba-inc.com, tpwu@1000video.com.hk}
}

\begin{document}
\maketitle

\begin{abstract}
To learn the optimal similarity function between probe and gallery images in Person re-identification, effective deep metric learning methods have been extensively explored to obtain discriminative feature embedding. However, existing metric loss like triplet loss and its variants always emphasize pair-wise relations but ignore the distribution context in feature space, leading to inconsistency and sub-optimal. In fact, the similarity of one pair not only decides the match of this pair, but also has potential impacts on other sample pairs. In this paper, we propose a novel Distribution Context Aware (DCA) loss based on triplet loss to combine both numerical similarity and relation similarity in feature space for better clustering. Extensive experiments on three benchmarks including Market-1501, DukeMTMC-reID and MSMT17, evidence the favorable performance of our method against the corresponding baseline and other state-of-the-art methods. 
\end{abstract}

\begin{IEEEkeywords}
Person Re-identification, Video Surveillance, Similarity Learning
\end{IEEEkeywords}

\section{Introduction}
Person re-identification (Re-ID) is a challenging problem. Assuming that there is no significant change in human appearance over a certain period of time, given a probe image of a person of interest, person images of same identity from a large gallery image database obtained by camera network can be searched and matched. It is an active research field and plays a critical role in surveillance scenarios. Person re-identification can be formulated as image retrieval problem where the distances between gallery images and the probe image are ranked according to the affinity matrix.

Convolutional neural networks (CNNs) based algorithms have played a dominant role in the field of Re-ID due to its power in learning discriminative and representative features. Deep learning related algorithms can be generally divided into three categories: (1) Sophisticated deep architecture design to extract robust and discriminative features. Among these algorithms, both global and local features have been utilized to explore hierarchical representative features \cite{DBLP:conf/eccv/SunZYTW18,DBLP:conf/cvpr/LiZG18,DBLP:conf/mm/WangYCLZ18}. (2) Deep metric learning algorithms \cite{shi2016embedding,hermans2017defense} are designed to minimize the intra-class divergence and to maximize the inter-class divergence at the same time. (3) The post-processing step \cite{bai2016sparse,zhong2017re} is normally excluded from the training phase, which aims at the affinity graph between gallery images to improve the initial rank list between probe image and gallery images. Some contextual information is utilized to adjust the relationships among images and boost the final matching performance, which make it convincing that contextual information is as important as pair-wise distance.

\begin{figure}[tbp]
\centerline{\includegraphics[width=8.7cm]{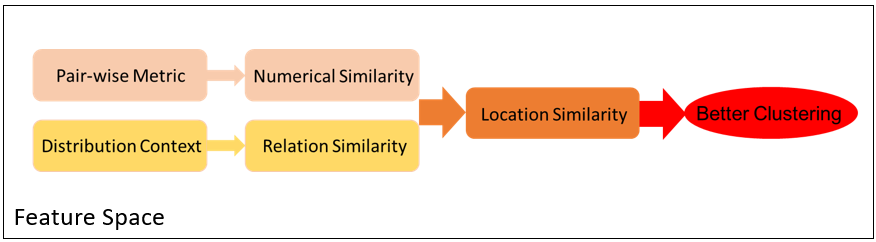}}
\setlength{\abovecaptionskip}{0pt}
\caption{The fusion of numerical similarity and relation similarity}
\label{fig}
\vspace{-0.6cm}
\end{figure}

In traditional deep metric learning based algorithms, pair-wise inputs are frequently involved in the widely adopted contrastive or triplet loss. In these settings, the feature distribution context from other instances is not fully explored to guide the parameter learning.  Generally, we assume that if the probe feature is close to a certain gallery feature in feature space, then they will have high probability of sharing the similar topological relationships with other samples in the feature space, which is called ``relation similarity" in this paper. Due to a lack to utilize the distribution context, ranking inconsistency commonly occurs among different probe-gallery pairs. As the figure 1 shows, contrastive and triplet loss can only describe the numerical similarity of features but ignore the contextual information encoded in the relationships among them. A distribution context descriptor should be designed to reflect the relation similarity. The integrated location similarity could be more accurately measured by the fusion of numerical similarity and relation similarity, which is very helpful for feature learning and spatial clustering.

This idea is also inspired by the core concept ``consistency" used in post-processing techniques, where relying on probe-to-gallery affinities to rank is not robust enough so that the gallery-to-gallery affinities are used to refine the rank list. In our case, that is, relying on pair-wise metric to measure features similarity is not robust enough so that we should fully explore the distribution context to refine the similarity in the learning phase. 

In this paper, we propose to incorporate distribution context into pair-wise metric to better describe features similarity in feature space by designing a simple but effective scheme. Then, we apply this method to two popular triplet losses, leading to better recognition performance. We name the refined loss as ``Distribution Context Aware" (DCA) loss. Finally, we make some modifications to the popular multiple granularity network \cite{DBLP:conf/mm/WangYCLZ18} to extract more fine-grained features, which can outperform most of existing state-of-the-art methods in recent years only with DCA loss.


Overall, the main contributions of this paper are summarized as follows:

\begin{list}{\labelitemi}{\leftmargin=1em}
    \setlength{\topmargin}{0pt}
    \setlength{\itemsep}{0em}
    \setlength{\parskip}{0pt}
    \setlength{\parsep}{0pt}
  \item We propose to explore the feature distribution context within a mini-batch to enhance the expressive power of pair-wise metric. Specifically, a kind of Distribution Context Aware (DCA) rule is utilized to refine the pair-wise metric. In this way, two popular forms of triplet loss are strengthened to learn more discriminative features.
  \item Extensive experiments on Market-1501 and DukeMTMC Re-ID datasets validate the effectiveness of the proposed DCA loss against the corresponding triplet loss baseline. 
  \item We made a modification to the popular multiple granularity network \cite{DBLP:conf/mm/WangYCLZ18} to extract more fine-grained feature merely with a DCA loss, which outperforms most existing state-of-the-art methods on three benchmarks including Market-1501, DukeMTMC-reID and MSMT17.
  \end{list}


\vspace{-0.2cm}
\section{Approach}
In this section, we introduce our method and network structure. Firstly we introduce the triplet Loss for metric learning briefly. Then, a detailed Distribution Context Aware (DCA) rule will be presented, following which the two triplet based forms of DCA loss are introduced. Finally, we will describe the overall network structure. 
\subsection{Triplet Loss for deep metric learning}
In deep metric learning, triplet loss is widely adopted to minimize the intra-person divergence while maximize the inter-person divergence. Specifically, given an image set $\mathcal{I}=\{\mathbf{I}_1,\mathbf{I}_2,\cdots,\mathbf{I}_N\}$ with $N$ images, we form the training mini-batch set into a set of triplets $\mathcal{T} = \{(\mathbf{I}_i,\mathbf{I}_j,\mathbf{I}_k)\}$, where $\mathbf{I}_i,\mathbf{I}_j,\mathbf{I}_k$ are images with identity labels $y_i$, $y_j$ and $y_k$ respectively. In a triplet unit, $(\mathbf{I}_i,\mathbf{I}_j)$ is a positive image pair of the same person (i.e., $y_i = y_j$), while $(\mathbf{I}_i,\mathbf{I}_k)$ is a negative image pair (i.e., $y_i\neq y_k$). Thus the purpose of Re-ID is to rank $\mathbf{I}_j$ before $\mathbf{I}_k$ for all triplets, which can be mathematically expressed as
\begin{equation}
d(\phi(\mathbf{I}_i),\phi(\mathbf{I}_j))+ \alpha \leq d(\phi(\mathbf{I}_i),\phi(\mathbf{I}_k)) \label{eq1}
\end{equation}
where $d(\mathbf{x},\mathbf{y})=\|\mathbf{x}-\mathbf{y}\|_2$ represents the Euclidean distance, $\phi(\cdot)$ denotes the feature transformation using deep neural networks as described later, and $\alpha>0$ is the margin by which the distance between a negative image pair is greater than that between a positive image pair. The whole loss function for all triplets in training mini-batch is then expressed as
\begin{equation}\small
\mathcal{L}_{tri}(\phi) = \frac{1}{|\mathcal{T}|} \sum_{(\mathbf{I}_i,\mathbf{I}_j,\mathbf{I}_k) \in \mathcal{T}} \Big[ d(\phi(\mathbf{I}_i),\phi(\mathbf{I}_j)) - d(\phi(\mathbf{I}_i),\phi(\mathbf{I}_k)) + \alpha \Big]_{+}
\label{eq2}
\end{equation}
where the operator $[\cdot]_{+}=\mathrm{max}(0,\cdot)$ represents the hinge loss as a common relaxation of Eq.~(\ref{eq1}) to enforce this constraint. And $|\mathcal{T}|$ denotes the number of triplets in $\mathcal{T}$.

Triplet loss represents a class of methods which is based on pair-wise distance calculation, comparison and ranking to distinguish the features difference belonging to varied identities. However, the distance can only reflect a certain norm of features numerical difference and may be confusing to a certain extent. It ignores the contextual information encoded in the relationships among other samples in mini-batch. 

\begin{figure}[tbp]
\vspace{-0.2cm}
\centerline{\includegraphics[width=8.7cm]{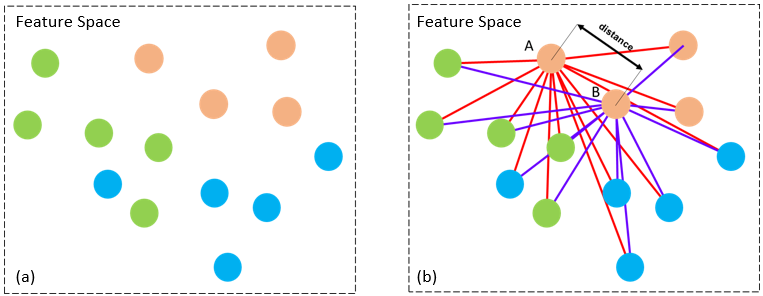}}
\setlength{\abovecaptionskip}{0pt}
\caption{(a) Feature Distribution among mini-batch images in feature space; (b) Distribution context of a pair of images, which reflects the relation similarity. The images with same identity should share similar spatial topological relations.}
\label{fig}
\vspace{-0.4cm}
\end{figure}

\subsection{Distribution Context Aware Rules}
In order to bring the distribution context into consideration, influences from other images in mini-batch are statistical analyzed to describe the similarity of spatial topological relations in feature space, as shown in figure 2 (b). The concept of exploring distribution context to refine the primary distance is commonly utilized in some algorithms \cite{bai2016sparse,ye2016person}. 

In set theory and some applications of data mining, the Jaccard distance \cite{levandowsky1971distance} is widely used to measure dissimilarity between sample sets. If we define the spatial relationship between a feature and other features as a set, Jaccard distance is able to describe the dissimilarity. In \cite{zhong2017re}, the Jaccard distance is utilized to re-rank the primary ranking list. 

We use the Gaussian kernel $V$ of the pair-wise distance to present the pair-wise similarity, which is calculated as:
\begin{equation}
V_{ij}=\mathit{e}^{-d(\phi(\mathbf{I}_i),\phi(\mathbf{I}_j))}
\label{eq5} 
\end{equation}
For a sample's feature $\mathbf{I}_i$, the similarity with other features in mini-batch constitutes a closed set, noted as $S\left [ \phi(\mathbf{I}_i)\right ]$. So the Jaccard distance between a pair of sets $S\left [ \phi(\mathbf{I}_i)\right ]$ and $S\left [ \phi(\mathbf{I}_j)\right ]$ can be mathematically formulated as 
\begin{equation}
  d_{Jaccard}(S\left [ \phi(\mathbf{I}_i)\right ], S\left [ \phi(\mathbf{I}_j)\right ])= 1 - \frac{\sum_{k=1}^{N}\min\left ( V_{ik},V_{jk} \right )}{\sum_{k=1}^{N}\max\left ( V_{ik},V_{jk} \right )}
    \label{eq6} 
\end{equation}
where $\mathbf{I}_i$ and $\mathbf{I}_j$ denote an image pair and $\mathbf{I}_k$ is other images in a mini-batch. 
Therefore, the Jaccard distance reflects the distribution compatibility of $\mathbf{I}_i$, $\mathbf{I}_j$ in feature space. Small Jaccard distance indicates that $\mathbf{I}_i$, $\mathbf{I}_j$ have similar topological relationships with other samples and dwell closer in feature space, and bigger Jaccard distance indicates a smaller overlapping among the distribution context of $\mathbf{I}_i$, $\mathbf{I}_j$. 

From the above analyses, by introducing Jaccard distance to re-weigh the primary feature distance, we can embed the distribution context into the primary metric loss. Therefore, the re-weighted distance can be calculated as
\begin{equation}\small
d_{weighted}(\phi(\mathbf{I}_i),\phi(\mathbf{I}_j))= d_{Jaccard}\ast  d(\phi(\mathbf{I}_i),\phi(\mathbf{I}_j))
\label{eq7} 
\end{equation}

The final distance $d_{DCA}$ is defined as:
\begin{equation}\small
d_{DCA}(\phi(\mathbf{I}_i),\phi(\mathbf{I}_j))= (1-\lambda)d+\lambda d_{Jaccard}+d_{weighted}  
\label{eq8} 
\end{equation}
where $\lambda\in [0,1]$ is the parameter to balance between the influence of different terms in Eq.~(\ref{eq8}) and encodes the distance to a new space.

\subsection{Batch Triplet-based Distribution Context Aware Loss}
We choose two popular forms of triplet loss as baselines to implement triplet-based DCA loss, which are Batch-Hard (\textbf{Tri-BH}) and Batch-All (\textbf{Tri-BA}) triplet loss \cite{hermans2017defense}. The mini-batch $\mathcal{T}$ is obtained by randomly sampling P identities, and then randomly sampling $K$ images of each identity, thus resulting in a batch of $PK$ images. The DCA distance in Eq.~(\ref{eq8}) is utilized to calculate the loss \textbf{DCA-BH} and \textbf{DCA-BA}:
\begin{equation}\small
\begin{split}
   \mathcal{L}_{DCA-BH}(\phi)=\frac{1}{|\mathcal{T}_{BH|}} \sum_{(\mathbf{I}_i,\mathbf{I}_j,\mathbf{I}_k) \in \mathcal{T}_{BH}} \Big[\max\limits_{j=1...M}d_{DCA}(\phi(\mathbf{I}_i),\phi(\mathbf{I}_j)) \\
   - \min \limits_{k=1...N}d_{DCA}(\phi(\mathbf{I}_i),\phi(\mathbf{I}_k)) + \alpha \Big]_{+} 
    \label{eq9} 
\end{split}
\end{equation}
\begin{equation}\small
\begin{split}
   \mathcal{L}_{DCA-BA}(\phi)=\frac{1}{|\mathcal{T}_{BA}|} \sum_{(\mathbf{I}_i,\mathbf{I}_j,\mathbf{I}_k) \in \mathcal{T}_{BA}} \Big[d_{DCA}(\phi(\mathbf{I}_i),\phi(\mathbf{I}_j)) \\
   - d_{DCA}(\phi(\mathbf{I}_i),\phi(\mathbf{I}_k)) + \alpha \Big]_{+} 
    \label{eq10} 
\end{split}
\end{equation}
where the M images are all the same ids and N images are all the negative ids in the batch. According to Eq.~(\ref{eq9})~(\ref{eq10}), the set $\mathcal{T}_{BH}$ and $\mathcal{T}_{BH}$ respectively contain $PK$ hard triplets and all possible $PK(PK-K)(K-1)$ combinations of triplets. During Tri-BA and DCA-BA loss training phrase, many of the possible triplets in a mini-batch are zero, essentially “washing out” the few significant contributing terms during averaging. We ran experiments where we only average the non-zero loss terms.

\subsection{Our CNN architecture}
The proposed architecture is shown in Figure 3. We use ResNet-50 \cite{he2016deep} as backbone network and multi-level horizontal stripes dividing method by Global Max Pooling (GMP) operations to obtain part-level features. The feature maps after $res\_conv4\_1$ block are then divided into two branches with different down-sampling strategies in $res\_conv5\_1$ block, sharing the similar architecture with ResNet-50 before $res\_conv5\_1$. The first branch adopts a stride-2 convolution layer as down-sampling while the second branch has no down-sampling operation. A GMP layer is then utilized to obtain the final features. Both the global and local features are then followed by a 1$\times$1 convolution layer with batch normalization and ReLU to reduce the featue to 256-dim. Then the multiscale features are concatenated together to capture the overall visual appearance. The triplet margin used is set to 1.2.

\section{EXPERIMENTS AND RESULTS}
\label{sec:pagestyle}

\subsection{Datasets}

\textbf{Market-1501}. Market-1501 is a popular available datasets for Re-ID with 32,668 annotated bounding boxes of 1501 subjects by 6 cameras. The dataset is split into 751 identities for training and 750 identities for testing.\\ 
\textbf{DukeMTMC-reID}. DukeMTMC-reID is collected with 8 cameras and used for cross-camera tracking. It contains
36,411 total bounding boxes from 1,404 identities, with half of them are used for training and the rest for testing.\\
\textbf{MSMT17}. MSMT17 is a larger and more challenging dataset, which captured by 13 outdoor cameras and 2 indoor cameras, and is collected under four different weather conditions. 4,101 identities and 126,441 bounding boxes are included.

\begin{figure}[tbp]
\vspace{-0.8cm}
\centerline{\includegraphics[width=8.7cm]{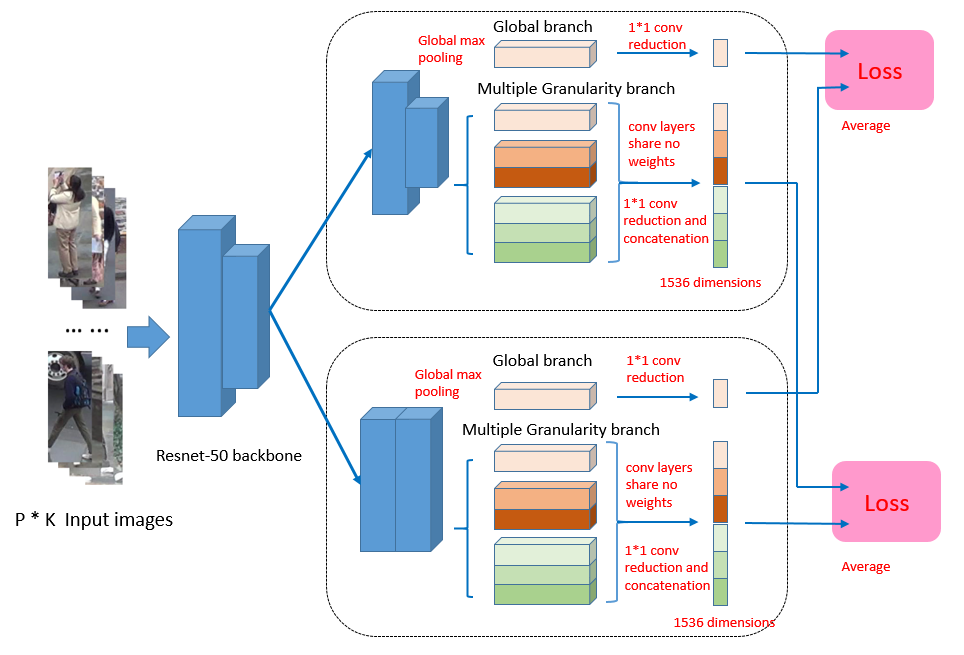}}
\setlength{\abovecaptionskip}{0pt}
\caption{Our CNN architecture.}
\label{fig}
\vspace{-0.6cm}
\end{figure}

\subsection{Implementation Details}
Our model is implemented on Pytorch framwork. The weights of pretrained ResNet-50 \cite{he2016deep} initializes the backbone and our branches before $res\_conv5\_1$ blocks. Input images are resized to 384$\times$128. $P=32, K=4$ as the batch-size. ADAM optimizer is used with momentum 0.9. We set $10^{-4}$ for initial learning rate. Learning rate decays to $10^{-5}$ and $10^{-6}$ after training for 220 and 320 epochs of overall 400 epoches. Random horizontal flipping for data augmentation and features averaged from original images and the horizontally flipped versions for evaluation are deployed. Resnet-50 \cite{he2016deep} as the baseline architecture in comparative study section, whose final layer is replaced by a fully-connected layer with 128 or 1024 output dimensions. $\lambda$ equals 0.5 and other settings are similar to the above. 

\begin{table}[th]
\begin{center}
\caption{
TABLE \RNum{1}: with baseline on dataset Market1501 and DukeMTMC-reID with different margins and output dimensions}
\label{table1}
\resizebox{9cm}{!}{
\begin{tabular}{r|cc|cc|cc|cc}
\hline
\multirow{3}{*}{Type-Margin} &  \multicolumn{4}{c|}{Market1501} & \multicolumn{4}{c}{DukeMTMC} \\
\cline{2-9}  & \multicolumn{2}{c|}{128dims} & \multicolumn{2}{c|}{1024dims} &  \multicolumn{2}{c|}{128dims} & \multicolumn{2}{c}{1024dims}  \\
\cline{2-9}& mAP & rank1 & mAP & rank1 & mAP & rank1 & mAP & rank1  \\
\hline
\noalign{\smallskip}
Tri-BH-0.5 &\textbf{64.1} &\textbf{82.4} &\textbf{64.9} &\textbf{83.0} &51.4 &69.4 &50.6 &\textbf{69.8}  \\
Tri-BH-0.8 &64.4 &81.9 &64.1 &81.9 &\textbf{52.1} &\textbf{70.8} &48.9 &67.9  \\
Tri-BH-1.2 &64.5 &82.0 &64.9 &82.5 &50.4 &69.4 &49.6 &68.8  \\
Tri-BA-0.5 &62.3 &80.9 &62.1 &81.7 &50.7 &66.9 &50.3 &67.5  \\
Tri-BA-0.8 &63.4 &80.9 &63.9 &81.9 &51.2 &68.9 &50.9 &66.9  \\
Tri-BA-1.2 &63.7 &81.6 &63.8 &82.3 &51.3 &69.8 &\textbf{51.3} &69.1  \\
\hline
DCA-BH-0.5 &\textbf{66.0} &\textbf{84.6} &\textbf{65.9} &84.2 &53.6 &72.3 &52.2 &\textbf{72.0}  \\
DCA-BH-0.8 &65.7 &84.1 &65.6 &84.3 &53.0 &71.9 &51.0 &71.3  \\
DCA-BH-1.2 &65.7 &84.3 &65.8 &\textbf{84.6} &\textbf{54.0} &\textbf{72.3} &\textbf{52.3} &71.5  \\
DCA-BA-0.5 &65.3 &83.9 &65.1 &83.1 &53.0 &72.1 &52.0 &70.9  \\
DCA-BA-0.8 &64.7 &82.7 &65.3 &83.6 &53.3 &71.9 &52.2 &71.3  \\
DCA-BA-1.2 &64.9 &82.5 &65.3 &83.5 &53.2 &72.2 &51.9 &71.6  \\
\hline
\end{tabular}}
\end{center}
\vspace{-0.3cm}
\end{table}

\begin{table}[th]
\begin{center}
\setlength{\belowcaptionskip}{10pt}
\caption{
TABLE \RNum{2}: Comparison with state-of-the-art methods on dataset Market1501 and DukeMTMC-reID
}
\label{table2}
\resizebox{9cm}{!}{
\begin{tabular}{r|c|cc|cc}
\hline
\multirow{2}{*}{Method} & \multirow{2}{*}{Reference} & \multicolumn{2}{c|}{Market1501} & \multicolumn{2}{c}{DukeMTMC} \\
\cline{3-6}     &     & mAP & rank1   & mAP & rank1 \\
\hline
\noalign{\smallskip}
AOS \cite{DBLP:conf/cvpr/HuangL0CH18} &CVPR2018 &70.4 &86.5 &62.1 &79.2\\
MLFN \cite{DBLP:conf/cvpr/ChangHX18} &CVPR2018 &74.3 &90.0 &62.8 &81.0\\
HA-CNN \cite{DBLP:conf/cvpr/LiZG18} &CVPR2018 &75.7 &91.2 &63.8 &80.5\\
GCSL \cite{chen2018group} &CVPR2018 &81.6 &93.5 &69.5 &84.9\\
PCB+RPP \cite{DBLP:conf/eccv/SunZYTW18} &ECCV2018 &81.6 &93.8 &69.2 &83.3\\
MGN \cite{DBLP:conf/mm/WangYCLZ18}&ACMMM2018 &86.9 &\textbf{95.7} &78.4 &\textbf{88.7}\\
MGN(DCA-BH) & &\textbf{87.1} &95.3 &\textbf{79.1} &\textbf{88.7}\\
\hline
DCA-BH+Res50  & &66.0 &84.6 &54.0 &72.3\\
DCA-BA+Res50  & &65.3 &83.9 &53.2 &72.2\\
Tri-BH+OurNet & &81.7 &92.1 &69.3 &82.1 \\
DCA-BH+OurNet-L0.5 & &83.4 &93.2 &\textbf{72.1} &84.1  \\
DCA-BH+OurNet-L0.8 & &\textbf{83.5} &\textbf{93.2} &71.9 &\textbf{84.5}  \\

\hline
\end{tabular}}
\end{center}
\end{table}

\begin{table}[t]
\vspace{-0.2cm}
\begin{center}
\caption{
TABLE \RNum{3}: Comparison with state-of-the-art methods on dataset MSMT17
}
\label{table3}
\begin{tabular}{lccc}
\hline\noalign{\smallskip}
Method &mAP & rank1 & rank5\\
\noalign{\smallskip}
\hline
\noalign{\smallskip}
GoogleNet \cite{DBLP:conf/cvpr/WeiZ0018} &23.0 &47.6 &65.0\\
PDC \cite{DBLP:conf/cvpr/WeiZ0018} &29.7 &58.0 &73.6\\
GLAD \cite{DBLP:conf/cvpr/WeiZ0018} &34.0 &61.4 &76.8\\
SFT \cite{DBLP:journals/corr/abs-1811-11405} &47.3 &\textbf{73.6} &86.0
\\\hline 
Tri-BH+OurNet &46.9 &71.4 &83.4 \\
DCA-BH+OurNet &\textbf{48.3} &72.0 &\textbf{86.1} \\
\hline
\end{tabular}
\end{center}
\vspace{-0.5cm}
\end{table}

\subsection{Comparison with baseline}
The detailed comparative study validates the impacts and effectiveness of DCA loss on Market-1501 and DukeMTMC datasets, compared with baseline triplet with different margins and output dimensions. Table \RNum{1} shows the results that the mean average precision (\textbf{mAP}) and \textbf{rank1} of DCA loss increase up to \textbf{1.9\%/2.2\%} for Batch-Hard and \textbf{3.0\%/3.0\%} for Batch-All in Market1501; and up to \textbf{3.6\%/2.9\%} for Batch-Hard and \textbf{2.3\%/5.2\%} for Batch-All in DukeMTMC. Evidences show that DCA-BA with more triplet samples to calculate and average has more improvement than DCA-BH, contrast with Tri-BA and Tri-BH respectively. This may be because distribution context makes statistics and averages throughout the batch, which can guide better clustering but cannot reach optimal minimum point as DCA-BH because of its over-averaged loss values. Overall, the DCA-BH still has better mAP and Rank1 accuracy than DCA-BA, but the performance gaps are very small. As we can see, different datasets with varied data distribution show different preferences for margin. And the performance of the model will not necessarily improve as the output dimension increases.

\vspace{-0.2cm}
\subsection{Comparison with State-of-the-art Methods}
To compare with the state-of-the-art methods, we choose some of the methods in recent top conference papers. Table \RNum{2}  presents the remarkable performance compared with other methods on Market1501 and DukeMTMC. Only with DCA-BH loss, our network could outperform most list state-of-the-art methods with \textbf{83.5\%/93.2\%} and \textbf{74.1\%/84.5\%} mAP/Rank1 in both Market1501 and DukeMTMC datasets, which also improve \textbf{2-3\%} mAP and \textbf{1-2\%} rank1 compared with our network with a Tri-BH loss.

Simultaneously, We also briefly compare the performance of different $\lambda$ values in DCA-BH loss in Table \RNum{2}, where $L0.8$ means $\lambda$ is set to 0.8. MGN is an extremely robust method and has excellent performance with the fusion of triplet and cross entropy. We tried to replace its triplet loss with our DCA-BH loss, and found its mAP still grows slightly in Table \RNum{2}. 

Finally, we trained and tested our model ($\lambda$ is set to 0.8) in the challenging large-scale dataset MSMT17, and the results show as Table \RNum{3}, our approach achieves an excellent performance, showing the strong adaptability and robustness in large complex scene. 
\vspace{-0.2cm}

\section{Conclusion}
\label{sec:typestyle}
 In this paper, we propose a novel distribution context aware loss combine both numerical similarity and relation similarity in feature space. Specifically, the Jaccard distance is utilized to adjust and encode the pair-wise distance. By taking the distribution context into account, it brings a notable performance gain to the traditional triplet loss baseline. In addition, with our network structure, we further achieve an excellent recognition performance. Extensive experiments on three large-scale benchmarks demonstrate the effectiveness of the proposed approach.
\vspace{-0.1cm}

\footnotesize
\bibliographystyle{IEEEbib}
\bibliography{czg}

\begin{thebibliography}{10}

\bibitem{DBLP:conf/eccv/SunZYTW18}
Yifan Sun, Liang Zheng, Yi~Yang, Qi~Tian, and Shengjin Wang,
\newblock ``Beyond part models: Person retrieval with refined part pooling (and
  {A} strong convolutional baseline),''
\newblock in {\em Computer Vision - {ECCV} 2018}.

\bibitem{DBLP:conf/cvpr/LiZG18}
Wei Li, Xiatian Zhu, and Shaogang Gong,
\newblock ``Harmonious attention network for person re-identification,''
\newblock in {\em 2018 {IEEE} Conference on Computer Vision and Pattern
  Recognition,}.

\bibitem{DBLP:conf/mm/WangYCLZ18}
Guanshuo Wang, Yufeng Yuan, Xiong Chen, Jiwei Li, and Xi~Zhou,
\newblock ``Learning discriminative features with multiple granularities for
  person re-identification,''
\newblock in {\em 2018 {ACM} Multimedia Conference on Multimedia Conference,
  {MM} 2018, Seoul, Republic of Korea, October 22-26, 2018}.

\bibitem{shi2016embedding}
Hailin Shi, Yang Yang, Xiangyu Zhu, Shengcai Liao, Zhen Lei, Weishi Zheng, and
  Stan~Z Li,
\newblock ``Embedding deep metric for person re-identification: A study against
  large variations,''
\newblock in {\em European Conference on Computer Vision}. Springer, 2016, pp.
  732--748.

\bibitem{hermans2017defense}
Alexander Hermans, Lucas Beyer, and Bastian Leibe,
\newblock ``In defense of the triplet loss for person re-identification,''
\newblock {\em arXiv preprint arXiv:1703.07737}, 2017.

\bibitem{bai2016sparse}
Song Bai and Xiang Bai,
\newblock ``Sparse contextual activation for efficient visual re-ranking,''
\newblock {\em IEEE Transactions on Image Processing}, vol. 25.

\bibitem{zhong2017re}
Zhun Zhong, Liang Zheng, Donglin Cao, and Shaozi Li,
\newblock ``Re-ranking person re-identification with k-reciprocal encoding,''
\newblock in {\em Computer Vision and Pattern Recognition (CVPR), 2017 IEEE
  Conference on}. IEEE.

\bibitem{ye2016person}
Mang Ye, Chao Liang, Yi~Yu, Zheng Wang, Qingming Leng, Chunxia Xiao, Jun Chen,
  and Ruimin Hu,
\newblock ``Person reidentification via ranking aggregation of similarity
  pulling and dissimilarity pushing,''
\newblock {\em IEEE Transactions on Multimedia}, vol. 18, no. 12, pp.
  2553--2566, 2016.

\bibitem{levandowsky1971distance}
Michael Levandowsky and David Winter,
\newblock ``Distance between sets,''
\newblock {\em Nature}, vol. 234, no. 5323, pp. 34, 1971.

\bibitem{he2016deep}
Kaiming He, Xiangyu Zhang, Shaoqing Ren, and Jian Sun,
\newblock ``Deep residual learning for image recognition,''
\newblock in {\em Proceedings of the IEEE conference on computer vision and
  pattern recognition}, 2016.

\bibitem{DBLP:conf/cvpr/HuangL0CH18}
Houjing Huang, Dangwei Li, Zhang Zhang, Xiaotang Chen, and Kaiqi Huang,
\newblock ``Adversarially occluded samples for person re-identification,''
\newblock in {\em 2018 {IEEE} Conference on Computer Vision and Pattern
  Recognition,}.

\bibitem{DBLP:conf/cvpr/ChangHX18}
Xiaobin Chang, Timothy~M. Hospedales, and Tao Xiang,
\newblock ``Multi-level factorisation net for person re-identification,''
\newblock in {\em 2018 {IEEE} Conference on Computer Vision and Pattern
  Recognition,}.

\bibitem{chen2018group}
Dapeng Chen, Dan Xu, Hongsheng Li, Nicu Sebe, and Xiaogang Wang,
\newblock ``Group consistent similarity learning via deep crf for person
  re-identification,''
\newblock in {\em Proceedings of the IEEE Conference on Computer Vision and
  Pattern Recognition}, 2018, pp. 8649--8658.

\bibitem{DBLP:conf/cvpr/WeiZ0018}
Longhui Wei, Shiliang Zhang, Wen Gao, and Qi~Tian,
\newblock ``Person transfer {GAN} to bridge domain gap for person
  re-identification,''
\newblock in {\em 2018 {IEEE} Conference on Computer Vision and Pattern
  Recognition, {CVPR} 2018, Salt Lake City, UT, USA, June 18-22, 2018}, 2018,
  pp. 79--88.

\bibitem{DBLP:journals/corr/abs-1811-11405}
Chuanchen Luo, Yuntao Chen, Naiyan Wang, and Zhaoxiang Zhang,
\newblock ``Spectral feature transformation for person re-identification,''
\newblock {\em CoRR}, vol. abs/1811.11405, 2018.

\end{thebibliography}

\vspace{12pt}

\end{document}